\documentclass[letterpaper]{article} % DO NOT CHANGE THIS
\usepackage{aaai23}  % DO NOT CHANGE THIS
\usepackage{times}  % DO NOT CHANGE THIS
\usepackage{helvet}  % DO NOT CHANGE THIS
\usepackage{courier}  % DO NOT CHANGE THIS
\usepackage{dsfont}
\usepackage{multirow}
\usepackage{amssymb}
\usepackage{amsmath}
\usepackage[hyphens]{url}  % DO NOT CHANGE THIS
\usepackage{graphicx} % DO NOT CHANGE THIS
\usepackage{natbib}  % DO NOT CHANGE THIS AND DO NOT ADD ANY OPTIONS TO IT
\usepackage{caption} % DO NOT CHANGE THIS AND DO NOT ADD ANY OPTIONS TO IT
\usepackage{algorithm}
\usepackage{algorithmic}

\usepackage{newfloat}
\usepackage{listings}
\DeclareCaptionStyle{ruled}{labelfont=normalfont,labelsep=colon,strut=off} % DO NOT CHANGE THIS
\floatstyle{ruled}
\newfloat{listing}{tb}{lst}{}
\floatname{listing}{Listing}
\title{Flexible 3D Lane Detection by Hierarchical Shape Matching}
\author{
    Zhihao Guan\textsuperscript{\rm 1},
    Ruixin Liu\textsuperscript{\rm 1},
    Zejian Yuan\textsuperscript{\rm 1},\\
    Ao Liu\textsuperscript{\rm 2},
    Kun Tang\textsuperscript{\rm 2},
    Tong Zhou\textsuperscript{\rm 2},
    Erlong Li\textsuperscript{\rm 2},
    Chao Zheng\textsuperscript{\rm 2},
    Shuqi Mei\textsuperscript{\rm 2}
}
\affiliations{
    \textsuperscript{\rm 1}Institute of Artificial Intelligence and Robotics, Xi’an Jiaotong University, China\\
    \textsuperscript{\rm 2}T Lab, Tencent Map, Tencent, China
    \{duduguan,sweetylrx\}@stu.xjtu.edu.cn, yuan.ze.jian@xjtu.edu.cn,\\\{allliu, kyriezhou, kunntang, erlongli, chrisczheng, shawnmei\}@tencent.com

}

\usepackage{bibentry}
\begin{document}

\maketitle

\begin{abstract}
As one of the basic while vital technologies for HD map construction, 3D lane detection is still an open problem due to varying visual conditions, complex typologies, and strict demands for precision. In this paper, an end-to-end flexible and hierarchical lane detector is proposed to precisely predict 3D lane lines from point clouds. Specifically, we design a hierarchical network predicting flexible representations of lane shapes at different levels, simultaneously collecting global instance semantics and avoiding local errors. In the global scope, we propose to regress parametric curves w.r.t adaptive axes that help to make more robust predictions towards complex scenes, while in the local vision the structure of lane segment is detected in each of the dynamic anchor cells sampled along the global predicted curves. Moreover, corresponding global and local shape matching losses and anchor cell generation strategies are designed. Experiments on two datasets show that we overwhelm current top methods under high precision standards, and full ablation studies also verify each part of our method. Our codes will be released at https://github.com/Doo-do/FHLD.
\end{abstract}

\section{Introduction}
While advanced driver-assistance system (ADAS) entering the lives of millions, one of its supporting tasks, centimeter-level lane detection for high-definition(HD) map construction is still limited by many \textbf{challenges}. First, diverse difficult visual conditions. The visual clue is not always clear in real driving scenes due to occlusions or severe illumination conditions. Second, complex typologies. The multifarious combination of varying numbers and directions of lanes, along with merges and splits make the discrimination of lane instances beset with difficulties. Moreover, more strict precision is needed. Widely used evaluation metrics like TuSimple’s are too permissive to errors in local offset\cite{polylanenet}, and thus methods with higher precision are still needed for accurate uses like HD map construction. 

Over the past decades, most researchers carry out lane detection with front-view images as input. Limited by the physical characteristics of RGB optical sensors, it cannot deal with poor illumination well. These years, more and more attention has been paid to lane detection with point cloud data. \textbf{Point cloud} is not only insensitive to illumination conditions, it can also preserve the accuracy of 3D information of lanes even in a very long distance. Moreover, its privacy-free feature is also a necessity for actual uses. Following all methods extracting lane lines from point cloud that we know, we structure the point cloud in a Bird's-eye-view (BEV) and regard encoded BEV feature maps as input.

Existing methods function in a segmentation manner which needs cumbersome clustering processes\cite{unet2}, or focuses on global information by predicting parametric lanes while ignoring local errors\cite{LSTR}, or some others based on preset anchors while losing the flexibility of predictions\cite{laneatt}.
In this paper, we propose an end-to-end framework with hierarchical outputs, to fully combine both the advantages of top-down manner which can collect all the global clues indicating instance information, and the strength of bottom-up manner that focuses on modeling local geometry and predicting accurate positions, to predict each lane as a sequence of points, providing more flexibility for detection results.

Specifically, in the global parametric curve branch, a Transformer structure is used to get a 3D parametric description for each lane. Notably, we propose a new representation for each curve based on the local adaptive reference axis, which incorporates the position of lane terminals, and represents the curve w.r.t its own reference axis to help the network fit better on complex scenes like oblique and curved lanes and hard topologies. Given the shape and instance information predicted from the global branch, dynamic anchor cells are generated along the lane and the corresponding cropped features are fed to the local branch for more precise and flexible predictions. 

In the local shape prediction branch, we detect line segment structure inside each anchor cell and use the center point of each as the final result, instead of directly detecting the specific lane center point inside every anchor. In addition, we optimize it by matching the ground truth Gaussian distribution along each lane segment, instead of regressing the segment parameters directly, introducing more explicit contextual information into account to shrink the searching space and optimize all the parameters simultaneously. Moreover, the positional relationships among predictions in different cells are also constrained, improving the smoothness of outputs. Notably, local predictions in dynamic anchor cells generated from global results are of certain instances and orders, thus eliminating clustering post-processing.

We collect a point cloud dataset RoadBEV on both highways and country roads in different cities for realistic evaluation. The results of comparison experiments on it and also on a self-annotated public dataset sub-KCUD, under a relatively strict metric with distance thresholds of 10 cm and 30 cm, show that our method outperforms recent state-of-the-art works in accurate detection. And also, full ablation studies validate the effectiveness of every part of our method. We summarise our main contributions as follows:

\begin{itemize}
\item An end-to-end hierarchical framework is designed which fully fuses both the global and local level information for accurate and flexible 3D lane detection.
\item We propose both a new global parametric representation for curves and a local shape description for line segments to enhance the robustness and flexibility of results.
\item The dynamic anchor cells generation module and also hierarchical shape matching strategy are proposed to help networks detect line elements hierarchically better.
\end{itemize}

\section{Related Works}
Traditional methods\cite{HSI, traditional1} exploit hand-crafted filters and adopt manual-designed processes like morphological filtering to extract and synthesize specialized features, resulting in  poor robustness. Recently, data-driven trends as represented by CNN make it easier to distinguish more subtle and complex features for lane detection. These deep learning methods can be divided into several paradigms as illustrated below.

\noindent\textbf{Segmentation-based methods.} Following the bottom-up manner, SAD\cite{enetsad} and RESA\cite{resa} predict a label for each pixel in the front-view camera images, and similar methods are also used for 3D point cloud data\cite{unet1, unet2}. These methods improve the detection performance by strengthening the priors of lanes or ameliorating networks for better feature extraction. However, cumbersome post-processing is still needed to cluster these isolated points into corresponding instances. Meanwhile, methods following this paradigm will output unnecessary points and be sensitive to local imperfections as well. By comparison, our method fully utilizes top-down information to guide the local shape prediction to generate flexible results and avoid post-processing.

\noindent\textbf{Grid-based methods.} This can be viewed as a combination of grid-level segmentation and accurate offset regression. PointLaneNet\cite{pointlanenet} directly predicts x-coordinates for fixed y values for each lane. UFast\cite{ufast} speeds up predictions by modeling it as a row-based selecting problem and handles the no-visual-clue problem by enlarging the receptive field. PINet\cite{pinet} improves the instance clustering by contrastively learning the embedding features for each grid. CondlaneNet\cite{condlanenet} resolves instance-level discrimination with the help of the proposal head and conditional convolution. The cells in these methods are not flexible so the performance may be influenced under the situations when two adjacent lanes fall in the same grid, or one lane line locates at the middle seam of two grids. To avoid these difficulties, our method predicts lane elements only within specific dynamically proposed anchor cells along the predicted global parametric curves.  

\noindent\textbf{Anchor-based methods.} Conventional detection methods\cite{dfpn} use the bounding box as a coarse proposal for the whole lane, which is not suitable for the slender structure of lanes. Dagmapper\cite{dagmapper} and HRAN\cite{hran} use iteratively proposed cells for local topology nodes searching from 3D point cloud data. However, they do not explicitly make full use of global semantic information, and thus local errors may be accumulated at the later stage of iterations. Some other methods like Lane-ATT\cite{laneatt} and SGNet\cite{structure} use dense line-shape anchors to provide better priors, where the anchors need to be pre-defined, and NMS is needed to remove redundant predictions. Following the idea of proposals but instead of setting dense anchors, we sample anchor cells along the predicted parametric lines to both provide a better global understanding and also enable end-to-end training. 

\noindent\textbf{Parametric-based methods.} LSTR\cite{LSTR}, PolyLaneNet\cite{polylanenet} and BezierLaneNet\cite{bezierlanenet} follow the top-down manner to model each lane as a parameterized curve and predict the global parameters directly from the network. However, these methods model the curve equation defined w.r.t fixed y-axis, which limits the flexibility for predicting diverse lanes. More importantly, local errors are ignored by these methods. In contrast, we propose a new 3D parametric representation for lane lines providing more flexibility, and also introduce extra local shape prediction to help get more precise localization.
\begin{figure*}[t] 
\centering 
\includegraphics[scale=0.625]{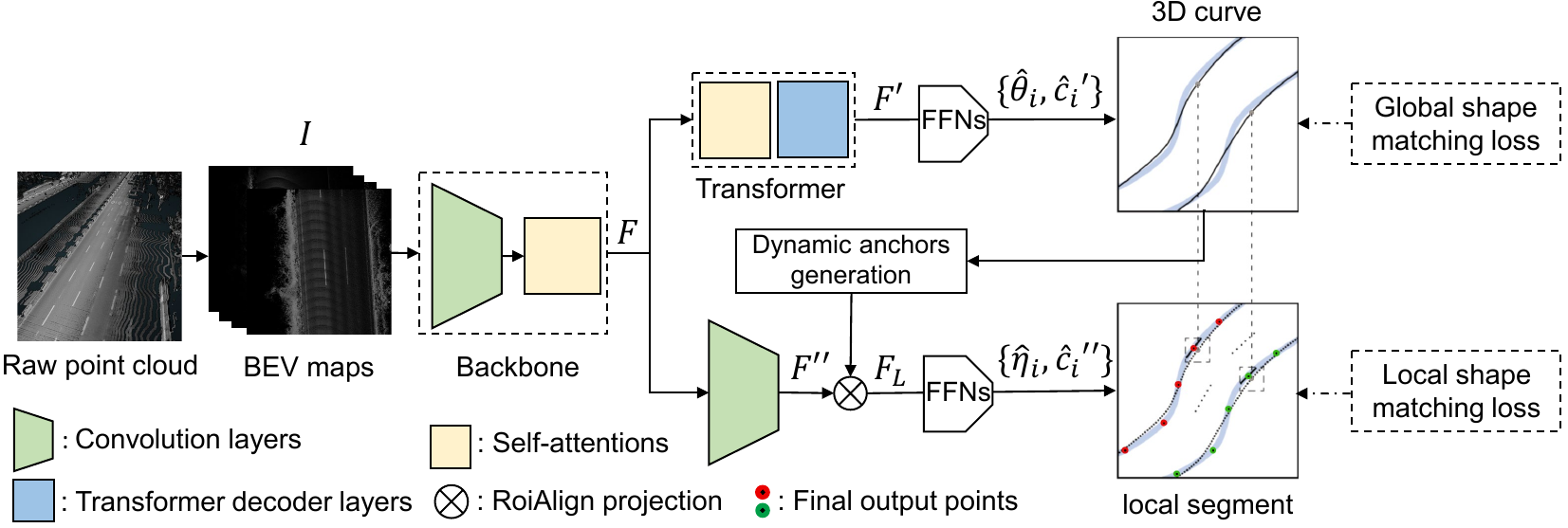} 
\caption{Framework structure. FHLD contains a shared backbone for feature extraction, two different branches separately detecting lane elements in different levels, and a dynamic anchors generation module to pass global prediction to local scope.} 
\label{fig:fig2} 
\end{figure*}
\section{Problem Definition}
We take the aggregated point cloud data as raw input, and feed the encoded BEV maps to networks to predict each lane line as a sequence of sparse 3D points along it, since the point set is more flexible for the slender structure of lane lines than other representations.

Since lane lines have few complex structures along height direction, we follow the common practice to structure the point cloud in BEV. Specifically, we project N points (each with 4 attributes: coordinates x, y, z, and reflective intensity r that distinguishes road markings from surroundings) onto the ground plane, rasterize it to $W\times H$ grids, and obtain the encoded BEV maps $I\in\mathds{R}^{W\times H\times4}$, of which the 4 channels in turn represent the average intensity, density, height variance, and minimum height of points in each grid.

\section{Methodology}
The end-to-end Flexible and Hierarchical Lane Detector (FHLD) we proposed is shown in Fig.\ref{fig:fig2}, where line elements are represented and predicted at different levels. For global understanding, each lane line is predicted in a parametric way based on its adaptive reference axis.
For more flexible and accurate outputs, a local line segment shape prediction branch is also proposed, to finally produce a sequence of points in dynamic anchor cells generated along each globally predicted curve. 
With accordingly designed hierarchical losses and training strategies, our method fully fuses high-level semantic information and local fine structure of lanes to both obtain global vision and avoid local errors.

\subsection{Hierarchical Shape Representation} 
\begin{figure}[t] 
\centering 
\includegraphics[scale=0.20]{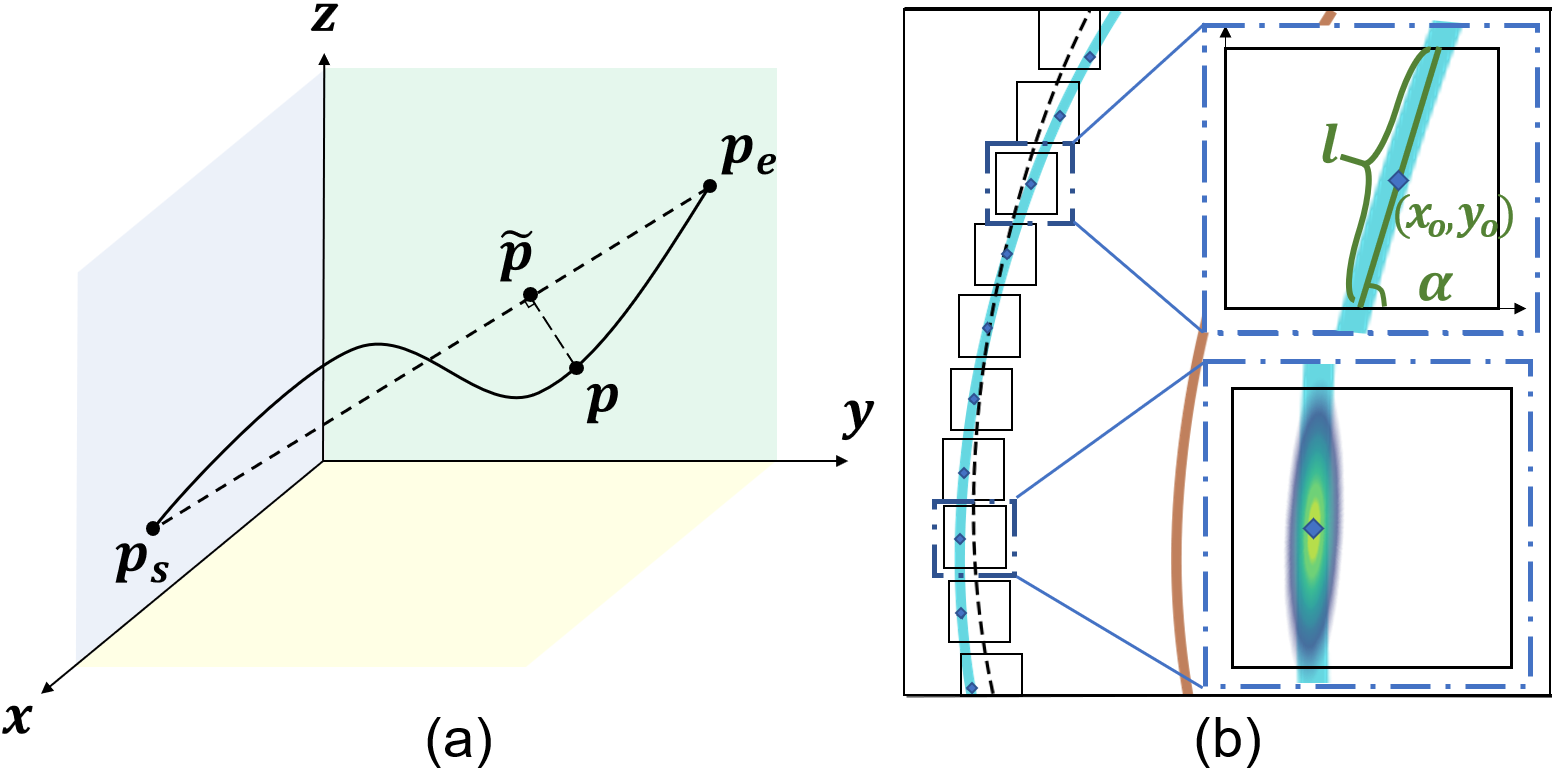} 
\caption{Hierarchical shape representation of lane lines. (a): Global parametric description. (b): Local segment shape representation \& Visualization of Gauss distribution.} 
\label{fig:fig3} 
\end{figure}

As shown in Fig.\ref{fig:fig3}, lane lines are represented at different levels, emphasizing global features of lane curves and local features of line segment shapes simultaneously to carry out hierarchical predictions.\\
\noindent\textbf{Global parametric shape representation.} To represent 3D lines in a more flexible way, we model each one as a cubic polynomial curve w.r.t. its own reference axis adaptively. Specifically, as shown in Fig.\ref{fig:fig3}(a), for each line we incorporate its starting position (lower terminal) $\boldsymbol{p_s}$ and ending position (higher one) $\boldsymbol{p_e}$ as a part of our parametric expression, which indicates the basic position of the lane. We set $\boldsymbol{p_s}$ as the origin and $\boldsymbol{\overrightarrow{p_sp_e}}$ as the positive direction of reference axis. Then, each line can be described with a set of curve parameters ${\theta}=(\boldsymbol{A}, \boldsymbol{B}, \boldsymbol{C}, \boldsymbol{D}, \boldsymbol{p_s}, \boldsymbol{p_e})$ 
which determines both the shape and position of this curve, written as:
\begin{equation}
\label{eq:global}
\boldsymbol{p} = f(t; {\theta})=\boldsymbol{A}t^3+\boldsymbol{B}t^2+\boldsymbol{C}t+\boldsymbol{D},
\end{equation}
\begin{equation}
\label{eq:global2}
t = \frac{||\overrightarrow{\boldsymbol{p_s\tilde{p}}}||}{||\overrightarrow{\boldsymbol{p_sp_e}}||},
\end{equation}
where $t\in[0, 1]$ denotes the relative position of any point $\boldsymbol{\tilde{p}}\in \mathds{R}^{3}$ moving on the reference axis between $\boldsymbol{p_s}$ and $\boldsymbol{p_e}$, $\boldsymbol{p}\in \mathds{R}^{3}$ denotes the corresponding point on the curve, and  $\boldsymbol{A}, \boldsymbol{B}, \boldsymbol{C}, \boldsymbol{D}\in \mathds{R}^{3}$ the cubic polynomial coefficients, $||\cdot||$ the magnitude of vector. 

Noted that when $t=0$ or $t=1$, corresponding reference point $\boldsymbol{\tilde{p}}$ overlaps on $\boldsymbol{p_s}$ or $\boldsymbol{p_e}$ respectively so the boundary conditions are shown below: 
\begin{equation}
\label{eq:border}
\left\{\begin{array}{l}
\boldsymbol{p_s}=f(0 ; {\theta})=\boldsymbol{D} \\
\boldsymbol{p_e}=f(1 ; {\theta})=\boldsymbol{A}+\boldsymbol{B}+\boldsymbol{C}+\boldsymbol{D}
\end{array}\right..
\end{equation}

It can be seen that 12 parameters (4 for each dimension) are required to describe each 3D curve. Compared with existing descriptions that model the curve w.r.t. y-axis, our formation gets rid of the binding with fixed reference axis, providing more flexibility and reasonable searching space, especially for inclined and complex lanes, as proven by experiments. Globally parametric prediction can fully excavate lane semantics in a top-down manner and provide abundant positional priors for local shape detection to further reduce local errors.

\noindent\textbf{Local shape description.}
To further predict more flexible and accurate results in local vision, each lane line is modeled as a sequence of segments, with their center points regarded as final outputs. Specifically, within dynamically sampled anchor cells along each lane (illustrated in the following section), we model the local line structure in each cell as a thin bar with a fixed pre-defined width $w_l$, as shown in Fig.\ref{fig:fig3}(b). Its position and shape can be described by a set of segment parameters ${\eta} = (\boldsymbol{p_o},{l}, {\alpha})$, where $\boldsymbol{p_o}$ denotes the 3D center point position of the bar, i.e. final outputs, $l$ is the length of lane segment in this cell, and $\alpha$ is the heading angle of this line segment. Notably, since the size of anchor cells only corresponds to a very tiny distance in real scenes, each comprising very little height difference, and thus the tilt and roll angles of line segments are just ignored in practice. 

Compared to directly modeling the local structure as the isolated center point, searching for a detailed bar structure helps to provide more context information in a smaller searching space, reducing the difficulty of regressing. Finally, with the curvature-aware dynamic anchoring strategy when inferencing, such local shape representation provides more flexible and locally accurate prediction results.
\subsection{Network Structure with Hierarchical Outputs}
FHLD contains a shared feature extraction backbone, two different branches separately searching lane elements at different levels, and a dynamic anchors generation module to pass global information for local segment shape searching. Specifically, input BEV maps $I$, our backbone composed of CNNs and a self-attention module fuses the global contextual information and outputs feature maps $F\in\mathds{R}^{w\times h\times c}$.

\noindent\textbf{Global parametric curve branch.} A DETR-like structure is adopted. To fully gather the context across entire feature maps to perceive the global shape, position, and instance information of lanes, extra self-attention blocks are adopted to encode $F$, and together with positional queries, they are then fed to Transformer decoder to generate $F'$, the attended feature sequences  for each of $N$ predictions. Finally, FFNs in the classification head and parameter head output result set $\{(\hat{c}’_i, \hat{\theta}_i)|i\text{=}1,..., N\}$, where $\hat{c}'_i$ are the confidence scores for each class (lane or non-lane). In practice, the outputs from each layer of the decoder are supervised.

\noindent\textbf{Local segments shape branch.}
After further encoding $F$ with convolution layers to $F''\in\mathds{R}^{w\times h\times c}$, 2D positions of generated anchors are mapped to $F''$ using RoiAlign\cite{maskrcnn} to crop corresponding feature sequences $F_L$. For each of these cropped local features, a set of results $\{(\hat{c}''_i, \hat{\eta}_i)|i\text{=}1, 2, ..., B\}$ is predicted, where $c''_i$ are predicted classification scores, $B$ is the number of cells generated on this lane. The position of segment center $\boldsymbol{\hat{p}_o}$ is predicted in a relative form w.r.t corresponding anchor cell center.

\noindent\textbf{Dynamic anchors generation
module.} Along each predicted parametric lane, a sequence of anchor cells is sampled following the strategy introduced in the following section. Every cell is projected onto $F''$ from the original resolution image and is quantified to $F_L$ with the help of RoiAlign. This projection guarantees that the local segment branch concentrates on the explicitly specific region of interest along the lane and reduces the false positive predictions.

\subsection{Training with Hierarchical Shape Matching Loss}
To supervise hierarchical outputs in two branches, The loss function is also designed in two parts accordingly.
\begin{equation}
\label{eq: vec}
L = L_{GSM} + L_{LSM}.
\end{equation}
\noindent\textbf{Global Shape Matching Loss.}
Since the ground truth lanes are described in the global coordinate system while the predicted ones are defined on the local reference axis, to define the difference between them, a bridge is needed. Specifically, we represent the $i$-$th$ ground truth lane $Q_i$ as a sequence of densely sampled points. For each point $\boldsymbol{p} \in Q_i$, its corresponding relative position $t_{p}$ on the reference axis can be calculated by projecting it to $\overrightarrow{\boldsymbol{p_sp_e}}$, written as:
\begin{equation}
\label{eq: vec}
t_{p} = \frac{\overrightarrow{\boldsymbol{p_sp_e}} \cdot\overrightarrow{\boldsymbol{p_sp}}}{||\overrightarrow{\boldsymbol{p_sp_e}}||^2},
\end{equation}
where $\cdot$ means dot product of two vectors, $\boldsymbol{p_s}$ and $\boldsymbol{p_e}$ are the ground truth terminals. At each relative position $t_{p}$, the corresponding prediction point $\boldsymbol{\hat{p}}=f(t_p, \hat{\theta_i})$ can be calculated. And then, a curve fitting loss $L_f$ is used to describe the gap between each pair of points on $i$-$th$ curve, written as:
\begin{equation}
\label{eq:Lossglobal2}
L_{f}(\hat{\theta}_{i}) =  \frac{1}{|Q_i|}\sum\limits_{\boldsymbol{p}\in Q_i}{{||\boldsymbol{\hat{p}}- \boldsymbol{p}||}} ,
\end{equation}
where $|\cdot|$ denotes the size of the set.

Moreover, a cross-entropy loss is used for the predicted logits as well. Overall, with $o({c_i})$ denoting the predicted probability for label ${c_i}\in\{0,1\}$, with assigning $\hat{\epsilon}(i)$-$th$ predicted lane matched to $Q_i$ after bipartite matching described in next section, and with ground truth set $Q$ padded to the number of N with non-lanes, the global shape matching loss can be written as:
\begin{equation}
\label{eq:Lossglobal}
L_{GSM}\!=\!\sum\limits_{i=1}\limits^{N}(\!-\lambda_{1}\! \log \!o_{\hat{\epsilon}(i)}\!({c}_{i})\!
+\!\mathds{1}\!({c}_{i}\neq0) L_{f}(\hat{\theta}_{\hat{\epsilon}(i)})), 
\end{equation}
where $\mathds{1}(\cdot)$ is an indicator function, and $\lambda_i$ is the weight to balance the effect of loss terms.

\noindent\textbf{Local Shape Matching Loss with Curve Smoothing.} Constraints on local branch are mainly composed of a shape matching loss $L_{kl}$ locating precise line segments, and a curve smoothing loss $L_{sm}$ aiming at constraining the smoothness of neighbored local predictions. With the binary classification loss written as $L_{cls}$ and $L_z$ denoting $L_2$ regression loss for the height of output points, it can be written as:
\begin{equation}
\label{eq:local}
\begin{aligned}
L_{LSM} = \lambda_2L_{kl} + \lambda_3L_{sm}
+ \lambda_1L_{cls}+ L_z.
\end{aligned}
\end{equation}

For better matching and locating the shape of segment elements in local scope, we optimize the Gauss distribution $g$ covered on each segment bar, determined by the parameter set $\eta$ (as visualized in Fig.\ref{fig:fig3}(b)). The mean $\mu$ and covariance $\Sigma$ of distribution $g$ can be written as:
\begin{equation}
\label{eq:gs2}
\begin{aligned}
\qquad \quad \boldsymbol{\Sigma}^{1 / 2}&=\mathbf{R} \Lambda \mathbf{R}^{\top}=\left(\begin{array}{cc}\cos \alpha & -\sin \alpha \\ \sin \alpha & \cos \alpha\end{array}\right)\cdot \\&\left(\begin{array}{cc}{w_l} & 0 \\ 0 & \frac{l}{2}\end{array}\right)\cdot\left(\begin{array}{cc}\cos \alpha & \sin \alpha \\ -\sin \alpha & \cos \alpha\end{array}\right),
\end{aligned}
\end{equation}
\begin{equation}
\label{eq:gs3}
\boldsymbol{\mu}=(x_o, y_o)^{\top},
\end{equation}
where $x_o$ and $y_o$ are the two components of $\boldsymbol{p_o}$, and $w_l$ is doubled to cover more contextual message.

Instead of directly regressing the isolated parameters of segments, we constrain the predicted distribution from the ground truth ones using KL-divergence, denoted as $L_{kl}$: 
\begin{equation}
\label{eq:gs4}
L_{kl} = \frac{1}{2}\sum\limits_{i=1}\limits^{N}\sum\limits_{j=1}\limits^{B}\left( KL(g_{i,j}|\hat{g}_{i,j})+KL(\hat{g}_{i,j}|{g_{i,j}})\right),
\end{equation}
where ${\hat{g}_{i,j}}$ is the predicted distribution in the $j$-$th$ anchor on $i$-$th$ lane. Such design makes its optimization easier than detecting a point or a rotated line directly because the position, angle, and length information of the slender structure can be optimized simultaneously with adaptively adjustable strategy, providing more accurate gradient information\cite{KLD}, and abundant contextual information around the lane is explicitly taken into account.

Moreover, thanks to the structured results of global branch, the anchor cells on each lane are of explicit order. We also constrain the positional relationship among predicted results in different cells. Concretely, $L_{sm}$ is used to constrain the second derivative of each curve, i.e., of consecutive output points, to guarantee the smoothness of predictions and avoid local imperfect predictions in unclear scenes like occlusion. With $\boldsymbol{\hat{p}_o^{i,j}}$ denoting the predicted absolute coordinate in the $j$-$th$ anchor on $i$-$th$ lane, it can be written as:
\begin{equation}
\label{eq:sm}
L_{sm}\!=\!\sum_{i=1}^{N}\!\sum_{j=1}^{B-2}\!\|(\boldsymbol{{\hat{p}_o^{i, j+2}}}-\boldsymbol{{\hat{p}_o^{i, j+1}}})\!-\!(\boldsymbol{{\hat{p}_o^{i, j+1}}}-\boldsymbol{{\hat{p}_o^{i, j}}})\|. 
\end{equation}
\subsection{Training Strategies}
\noindent\textbf{Dynamic anchors generation.}
The local structure of the lane line is searched under the guidance of top-down information, obtaining a better understanding of the entirety. Say concretely, a sequence of anchor cells is sampled dynamically along each parametric lane for further local fine prediction. Dynamic anchoring, instead of conventional fixed grid-based methods, can avoid both dense computations and failure in troublesome situations when the line lies at the middle seam of two grids, or two lanes are in the same grid. 

When training, $B$ points are sampled along each predicted parametric lane as the centers of dynamic anchor cells. According to empirical attempts and experimental results, the shape and size of cells here do not affect the results much so we set a fixed size for anchors at $r\times r$ pixels. Thanks to the global branch, here most of the sampled cells contain lane segments, i.e., are positive samples. For better distinguishing the local features, following the idea of contrast learning, negative anchor cells are also randomly sampled around each positive one at a set proportion. Notably, only classification loss $L_{cls}$ is used to supervise these negative samples. Moreover, a little random offset is added as noise to the sampled positive positions to mimic the imperfection in global predictions and improve the generalization.

When inferencing, only positive anchors are sampled, and the sampling rate of each lane is set proportional to its curvature. That is to say, there will be more points predicted on the complex lane structure than on the simple straight one, keeping the compactness of outputs. 

\noindent\textbf{Bipartite matching for global predictions.}
The correspondence between $N$ predictions and $M$ (padded to N) ground truth lanes is built using Hungarian Algorithm (HA) following LSTR. To design the cost matrix $D\in \mathds{R}^{N\times N}$, only two parts are covered in it, the classification cost and the position cost. Though a more precise comparison between two line instances can be done, under our adaptive parametric representation of curves it is sufficient to match the terminals due to their unambiguity and importance in all road scenes. With $d_{i,j}$ denoting the $(i$-$th, j$-$th)$ element in $D$, and $o_j({c_i})$ the probability of target class label ${c_i}$ predicted by $j$-$th$ network output, it can be written as:
\begin{equation}
\label{eq:d_ij}
d_{i,j}\!=\!-\!\lambda_{1}\! o_j\!(c_i)+\mathds{1}\!\left(\!c_i\!\neq\!0\!\right)\!  L_1\!\left((\boldsymbol{\hat{p}_{s}^{(\!j\!)\!}},\boldsymbol{\hat{p}_{e}^{(\!j\!)\!}}),  (\boldsymbol{p_{s}^{(\!i\!)\!}},  \boldsymbol{p_{e}^{(\!i\!)\!}})\right).
\end{equation}

With the help of HA, a one-to-one match can be guaranteed where the $i$-$th$ ground truth lanes are matched with the $\hat{\epsilon}(i)$-$th$ prediction with $\hat{\epsilon}=\mathop{\arg\min}\limits_{\epsilon}\sum_{i=1}^{N}d_{i,\epsilon(i)}$.

\section{Experiments}
To show the superiority of our framework and also the effectiveness of each design, we have done abundant comparative experiments and ablation studies.

\noindent\textbf{Dataset:} Experiments are carried out on two point cloud dataset, a self-collected one named RoadBEV and sub-KCUD, a subset of public KAIST Complex Urban Dataset \cite{keist} annotated by ourselves, since there are no readily available large-scale public datasets with 3D lane annotations for the HD map construction task. For each dataset, from the raw point cloud, regions of 25$\times$25 m are cropped along the trajectory at a stride of 13 m. Then we project and rasterize the points to 800$\times$800 pixel BEV maps encoded as mentioned before. To annotate each lane, 3D polylines crossing the center of lane lines are used. At most 8 and at least 0 lines are contained in each BEV map, and complex scenes like splits and merges are included. RoadBEV contains around 25k and sub-KCUD contains nearly 4k BEV images. For each dataset, we randomly set 80\% sequences for training and 20\% for testing. 

\noindent\textbf{Metrics:} Following the testing paradigm of Dagmapper, we set a more strict evaluation scheme compared with the ones defined by Tusimple or Culane, which is more necessary for high-precision HD maps. Firstly, a large number of points are densely sampled respectively from predicted and ground truth lanes, and then, recall and precision rates are calculated by searching the number of ground truth or predicted points within a distance threshold from the predicted or ground truth points. Under the 10 cm and 30 cm distance threshold, we report the precision(\%), recall(\%), F1 score(\%), and also FPS of all methods. Notably, we care more about stricter \textbf{10 cm} standard. FPS is calculated with batch size 1 on Nvidia RTX3090 implemented by Pytorch. 

\begin{table*}[ht]
\setlength{\tabcolsep}{1.12mm}

\begin{tabular}{|l|cccccc|cccccc|c|}
\hline
\multicolumn{1}{|c|}{\multirow{3}{*}{Method}} & \multicolumn{6}{c|}{RoadBEV}                                                                                                                              & \multicolumn{6}{c|}{sub-KCUD} 
&\multicolumn{1}{|c|}{\multirow{3}{*}{FPS}}
\\ \cline{2-13} 
\multicolumn{1}{|c|}{}                        & \multicolumn{3}{c|}{\textbf{10 cm}}                                                             & \multicolumn{3}{c|}{30 cm}                                        & \multicolumn{3}{c|}{\textbf{10 cm}}                                                             & \multicolumn{3}{c|}{30 cm}                                      &  \\ \cline{2-13} 
\multicolumn{1}{|c|}{}                        & \multicolumn{1}{c|}{Precision} & \multicolumn{1}{c|}{Recall} & \multicolumn{1}{c|}{F1} & \multicolumn{1}{c|}{Precision} & \multicolumn{1}{c|}{Recall} & F1 & \multicolumn{1}{c|}{Precision} & \multicolumn{1}{c|}{Recall} & \multicolumn{1}{c|}{F1} & \multicolumn{1}{c|}{Precision} & \multicolumn{1}{c|}{Recall} & F1 &\\ \hline
LaneATT                                       & \multicolumn{1}{c|}{71.72}          & \multicolumn{1}{c|}{70.35}       & \multicolumn{1}{c|}{71.03}   & \multicolumn{1}{c|}{90.26}          & \multicolumn{1}{c|}{89.99}       & 90.12   &  \multicolumn{1}{c|}{72.15}          & \multicolumn{1}{c|}{71.99}       & \multicolumn{1}{c|}{72.07}   & \multicolumn{1}{c|}{91.32}          & \multicolumn{1}{c|}{90.02}   & 90.67  &94    \\
LSTR                                          & \multicolumn{1}{c|}{71.70}          & \multicolumn{1}{c|}{72.24}       &   \multicolumn{1}{c|}{71.97}   & \multicolumn{1}{c|}{91.51}          & \multicolumn{1}{c|}{90.23}       &  90.87  & \multicolumn{1}{c|}{72.01}          & \multicolumn{1}{c|}{72.40}       & \multicolumn{1}{c|}{72.20}   & \multicolumn{1}{c|}{91.45}          & \multicolumn{1}{c|}{90.80}       &91.12   &116 \\
PINet                                         & \multicolumn{1}{c|}{70.92}          & \multicolumn{1}{c|}{66.32}       & \multicolumn{1}{c|}{68.54}   & \multicolumn{1}{c|}{92.33}          & \multicolumn{1}{c|}{90.27}       & 91.29   & \multicolumn{1}{c|}{71.58}          & \multicolumn{1}{c|}{67.85}       & \multicolumn{1}{c|}{69.67}   & \multicolumn{1}{c|}{90.39}          & \multicolumn{1}{c|}{88.99}       &  89.68   &5\\
UFast                                         & \multicolumn{1}{c|}{63.52}          & \multicolumn{1}{c|}{68.34}       & \multicolumn{1}{c|}{65.84}   & \multicolumn{1}{c|}{87.55}          & \multicolumn{1}{c|}{93.81}       & 90.57   & \multicolumn{1}{c|}{69.92}          & \multicolumn{1}{c|}{68.16}       & \multicolumn{1}{c|}{69.03}   & \multicolumn{1}{c|}{89.02}          & \multicolumn{1}{c|}{\textbf{93.12}}       & 91.02  &\textbf{133} \\
CondLaneNet                                   & \multicolumn{1}{c|}{74.13}          & \multicolumn{1}{c|}{73.02}       & \multicolumn{1}{c|}{73.57}   & \multicolumn{1}{c|}{\textbf{93.09}}          & \multicolumn{1}{c|}{92.97}       &  93.03  & \multicolumn{1}{c|}{72.25}          & \multicolumn{1}{c|}{72.30}       & \multicolumn{1}{c|}{72.27}   & \multicolumn{1}{c|}{90.30}          & \multicolumn{1}{c|}{92.79}       &  91.53 &75 \\
SAD                                      & \multicolumn{1}{c|}{71.33}          & \multicolumn{1}{c|}{70.25}       & \multicolumn{1}{c|}{70.79}   & \multicolumn{1}{c|}{89.77}          & \multicolumn{1}{c|}{93.42}       & 91.56   & \multicolumn{1}{c|}{68.72}          & \multicolumn{1}{c|}{67.99}       & \multicolumn{1}{c|}{68.35}   & \multicolumn{1}{c|}{89.71}          & \multicolumn{1}{c|}{90.89}       & 90.30  & 35\\
PolyLaneNet                                   & \multicolumn{1}{c|}{58.69}          & \multicolumn{1}{c|}{57.07}       & \multicolumn{1}{c|}{57.87}   & \multicolumn{1}{c|}{82.23}          & \multicolumn{1}{c|}{80.99}       & 81.61   & \multicolumn{1}{c|}{60.73}          & \multicolumn{1}{c|}{59.78}       & \multicolumn{1}{c|}{60.25}   & \multicolumn{1}{c|}{83.25}          & \multicolumn{1}{c|}{84.39}       & 83.82 & 104 \\
RESA                                   & \multicolumn{1}{c|}{60.16}          & \multicolumn{1}{c|}{56.45}       & \multicolumn{1}{c|}{58.25}   & \multicolumn{1}{c|}{79.80}          & \multicolumn{1}{c|}{80.75}       & 80.27   &  \multicolumn{1}{c|}{62.93}          & \multicolumn{1}{c|}{60.42}       & \multicolumn{1}{c|}{61.65}   & \multicolumn{1}{c|}{80.65}          & \multicolumn{1}{c|}{78.32}       &  79.47 & 25\\
 \hline \hline
FHLD(Ours)                                    & \multicolumn{1}{c|}{\textbf{76.68}}          & \multicolumn{1}{c|}{\textbf{78.49}}       & \multicolumn{1}{c|}{\textbf{77.57}}   & \multicolumn{1}{c|}{92.99}          & \multicolumn{1}{c|}{\textbf{94.50}}       & \textbf{93.74}   & \multicolumn{1}{c|}{\textbf{74.42}    }      & \multicolumn{1}{c|}{\textbf{74.45}}       & \multicolumn{1}{c|}{\textbf{74.43}}   & \multicolumn{1}{c|}{\textbf{91.49}}          & \multicolumn{1}{c|}{93.05}       & \textbf{92.20}  &97 \\ \hline
\end{tabular}

\caption{Comparison experiments results on RoadBEV and sub-KCUD dataset.}
\label{tab:comp1}
\end{table*}

\noindent\textbf{Details:} For backbone, we modify commonly used Resnet-18\cite{resnet} by cutting the output channels of each block into [16, 32, 64, 128] to avoid overfitting. We augment each input BEV map by random shifting, rotating, flipping, scaling, and cropping, transforming it to a size of $640\times640$. We train each model for 200k iterations with batch size 64 and initial learning rate 0.0001 using Adam optimizer. The loss coefficients $\lambda_1$, $\lambda_2$ and $\lambda_3$ are set to 1, 1 and 0.5, anchors sampling ratio is 1:100, and $N$=15, $B$=40. If multiple segments appeared in one cell, the one closest to the cell center is chosen as the ground truth in this cell. For better performance, $L_{LSM}$ is not optimized for first 2k iterations. All of the backbones and hyper-parameter settings are set to the same for fair comparison for all experiments if appliable. Three parallel experiments are carried out for each method. 
\begin{table*}[t]

\setlength{\tabcolsep}{1.71mm}
\begin{tabular}{|cc|cc|cc|ccc|ccc|}
\hline
\multicolumn{2}{|c|}{Global description}                    & \multicolumn{2}{c|}{Local matching}                    & \multicolumn{2}{c|}{Training}                       & \multicolumn{3}{c|}{\textbf{Results (10 cm)}}     & \multicolumn{3}{c|}{Results (30 cm)}                                                   \\ \hline
\multicolumn{1}{|c|}{\enspace Adp. \quad} &Fixed                   & \multicolumn{1}{c|}{Shape} & Point                     & \multicolumn{1}{c|}{$L_{sm}$} & Noise             & \multicolumn{1}{c|}{Precision} & \multicolumn{1}{c|}{Recall} & F1& \multicolumn{1}{c|}{Precision} & \multicolumn{1}{c|}{Recall} & F1 \\ \hline
\checkmark       &                           & \checkmark  &                           & \checkmark        & \checkmark &  76.68                                 &           78.49                 &  77.57 &92.99&94.50&93.74  \\\hline
                                & \checkmark & \checkmark  &                           & \checkmark        & \checkmark & 74.89                              &76.93                             & 75.90(-1.67) &92.47&94.01&93.23(-0.51)  \\
\checkmark       &                           &                            & \checkmark & \checkmark        & \checkmark &               74.77                 &  76.48                           & 75.62(-1.95) &92.36&93.84&93.09(-0.65)  \\ \hline
\checkmark       &                           &                            &                           &                                  &                           &    72.66                            &  73.25                           &  72.95(-4.62) &92.18&91.63&91.99(-1.75) \\
      &             \checkmark               &                            &  &         &  &  71.73                                &   72.23                          & 71.98(-5.59)&91.50&90.24&90.87(-2.87) \\ \hline
                                
\checkmark       &                           & \checkmark  &                           &                                  & \checkmark &  75.58                              &                      77.79       & 76.67(-0.90) &92.49&94.22&93.35(-0.39)   \\
\checkmark       &                           & \checkmark  &                           & \checkmark        &                           &   75.50                             &               77.81              &76.64(-0.93) &92.53&94.25&93.38(-0.36)      \\ \hline
\end{tabular}
\caption{Ablation studies results on RoadBEV dataset.}
\label{tab:ablation}

\end{table*}
\subsection{Comparative Experiments}
\noindent\textbf{Baseline:} Because there are few strong open-sourced baselines for point cloud lane detection, and due to its similarity to front-view detection tasks, we have to modify some strongest RGB methods to adapt to this task as comparisons. Our method is compared with adapted Lane-ATT, LSTR, PINet, UFast, CondLaneNet, SAD, PolyLaneNet, and RESA. Notably, since most of these methods are hard to expand to 3D prediction, we directly set their height predictions as the ground truth ones, while ours predicts 3D coordinates straight from the model outputs. That is to say, we put other methods we compare with in a better position.

It's important to note that SAD, RESA, and UFast all need to pre-define the instance labels, and we assign each lane one of 10 labels according to the horizontal positions. LSTR's modeling for curves is designed for the geometry of front-view images so we adapt it to polynomial curves in BEV. Methods beside LSTR need more or fewer post-processings and lane-ATT needs pre-designed anchor settings. In contrast, our method trains and inferences in a totally end-to-end manner, regardless of the number, position and direction of the lanes in the image. Besides, our method can flexibly output point set at different sparsities depending on the curvature, or following other customized requirements. In this section the qualitative results of output point set are shown equidistantly for clear comparison with other methods. More flexible output options demonstrated in 3D point clouds, as well as more hard samples of complex topologies with both intermediate and final results are visualized in the \textbf{Supp.}. 

\begin{figure*}[t] 
\centering 
\includegraphics[scale=0.385]{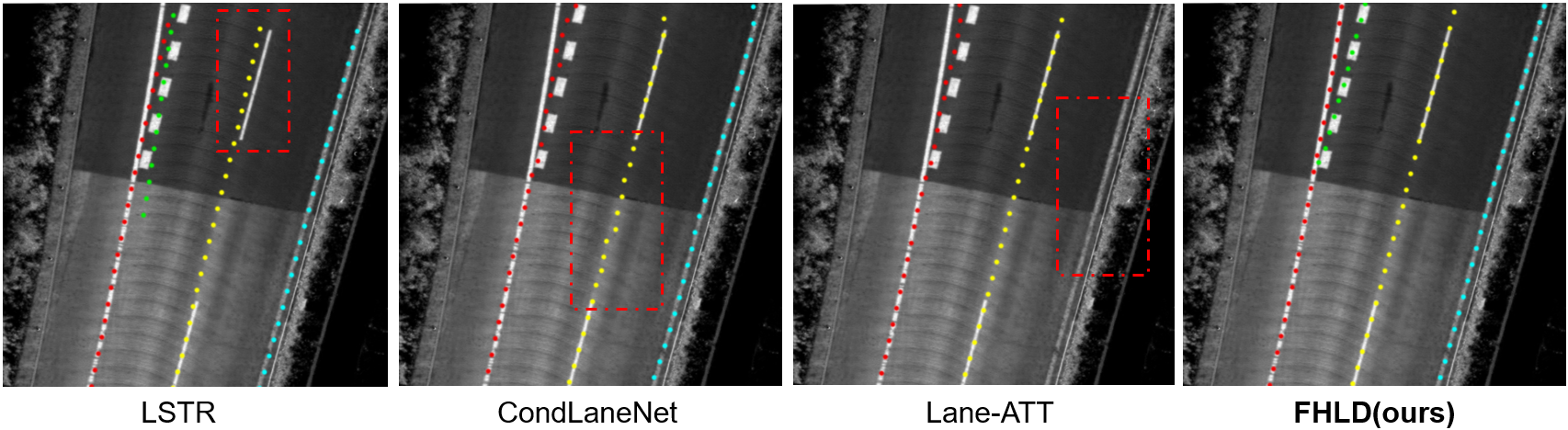} 
\caption{Visualization of prediction results on RoadBEV (Best viewed in colors).} 
\label{fig:vis}
\end{figure*}

The testing results on two datasets can be found in Tab.\ref{tab:comp1}. Clearly, although our method is tested under a harder situation, it still overwhelms others, especially under the more accurate 10 cm standard since our flexible output can greatly improve local errors. Specifically, on RoadBEV dataset, FHLD respectively obtains 5.4\% and 3.0\% higher F1 score than the second best-performed method under 10 cm distance threshold, and obtains 0.8\% and 0.7\% higher under 30 cm distance threshold. Our methods successfully improve the accurate prediction at the centimeter  level. Compared with the parametric-based and cell-based methods most similar to ours, LSTR and PINet, we achieve 7.8\% and 13.2\% higher scores under 10 cm standard on RoadBEV. Besides, we can notice that thanks to the anchor generation module, the recall of FHLD is also improved much, especially when compared to segmentation methods. In addition, The corresponding qualitative results can be seen in Fig.\ref{fig:vis}, where FHLD performs accurately thanks to flexible and hierarchical outputs, while others either ignore local offsets(LSTR), lack constraints on relative relationships(CondLaneNet), or just miss the whole line due to preset anchors(Lane-ATT).
\subsection{Ablation Studies}
To show the contributions of each part in our method, we carry out abundant ablation studies, including replacing our global curve representation w.r.t adaptive axis(\textbf{Adp.}) with traditional description w.r.t fixed axis(\textbf{Fixed}), replacing local shape Gauss matching(\textbf{Shape}) with direct lane center point regression(\textbf{Point}), with or without $\boldsymbol{L_{sm}}$ and \textbf{noise} when generating anchors. The results of each combination are shown in Tab.\ref{tab:ablation}, where we can see that the local prediction branch contributes much to the final result (+4.62) and the new parametric representation of the global curve (+1.67) as well as the local shape matching strategy (+1.95) also helps a lot, especially under 10 cm threshold.
\begin{figure*}[h] 
\centering 
\includegraphics[scale=0.3725]{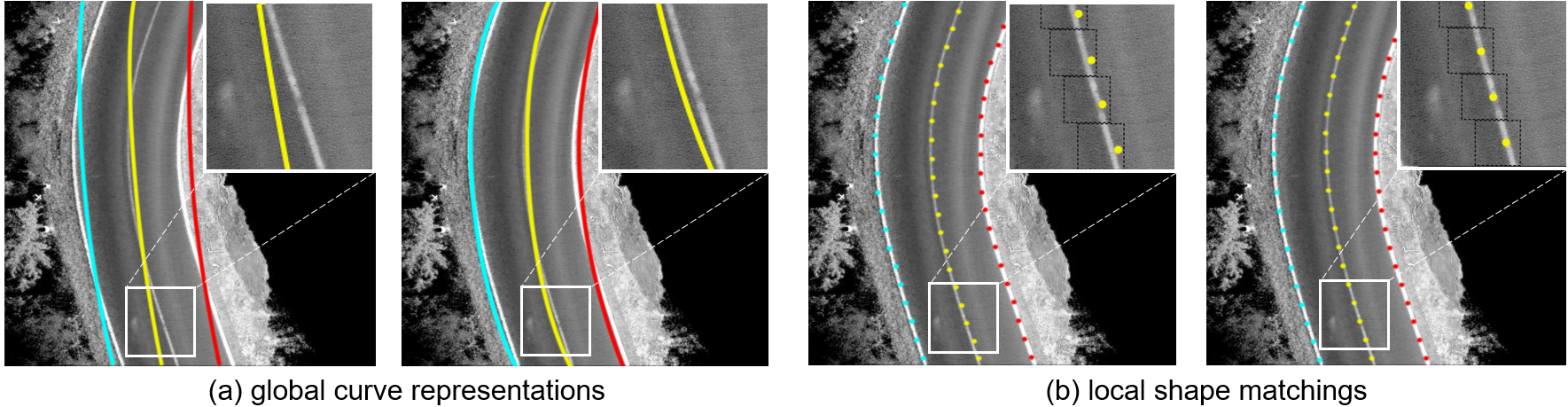} 
\caption{Prediction results visualization of ablation studies (Best viewed in colors). (a): Direct global output with fixed(left) and adaptive(right) representations. (b): Final point output of local prediction by point regressing(left) or shape matching(right).} 
\label{fig:abl2} 
\end{figure*}

\noindent\textbf{Effectiveness of global curve representation.} To better show the superiority of our global representation, we further test our method with adaptive(\textbf{Adp.}) or traditional(\textbf{Fixed}) representations on a subset dataset hard-RoadBEV, which comprises only lanes with large scopes and curvatures. We report both the results of parametric curves (without local branch) and final point outputs. The results shown in Tab.\ref{tab:abl1} validate that modeling curves under the adaptive axis helps to fit such scoped curves better than the original ones. Additionally, though traditional representation result is 6.2\% lower than our representation, after the local predictions, the gap is narrowed to 4.0\%, showing the advantage of local branch in helping with imperfect curve fittings for robust output. Corresponding visualization in Fig.\ref{fig:abl2}(a) also shows the validation of our global representation on flexibility. %to fit on complex road scenes accurately.

\noindent\textbf{Contributions of local shape prediction and matching.} Tab.\ref{tab:ablation} shows that without local shape prediction, directly outputting global curves will lower F1 by 6.0\%. Changing from local shape matching to local point regressing lower F1 by 2.5\%. Also, Fig.\ref{fig:abl2} shows that our local shape matching relieves the influence brought by imperfect global prediction, and avoids local errors under high precision requirements. 
\begin{table}[t]
\centering 

\setlength{\tabcolsep}{0.8mm}
\begin{tabular}{|c|ccc|ccc|}
\hline
\multirow{2}{*}{} & \multicolumn{3}{c|}{Parametric output}                            & \multicolumn{3}{c|}{Point set output}                            \\ \cline{2-7} 
                  & \multicolumn{1}{c|}{Precision} & \multicolumn{1}{c|}{Recall} & F1 & \multicolumn{1}{c|}{Precision} & \multicolumn{1}{c|}{Recall} & F1 \\ \hline
                  Adp.& \multicolumn{1}{c|}{71.07} & \multicolumn{1}{c|}{71.65} & 71.36 &  \multicolumn{1}{c|}{74.62} &  \multicolumn{1}{c|}{74.73} & 74.67  \\ \hline
                 Fixed&
                 \multicolumn{1}{c|}{67.24} & \multicolumn{1}{c|}{66.58} &  66.91 & \multicolumn{1}{c|}{71.35} &  \multicolumn{1}{c|}{72.01} & 71.68 \\ \hline
\end{tabular}
\caption{Results on hard-RoadBEV under 10 cm standard.}
\label{tab:abl1}

\end{table}
\begin{table}[t]
\centering 

\setlength{\tabcolsep}{3mm}
\begin{tabular}{|c|c|c|c|c|c|}
\hline
r  & 16    & 24    & 32             & 40    & 48    \\ \hline
F1 & 77.23 & 77.45 & \textbf{77.57} & 77.47 & 77.36 \\ \hline
\end{tabular}
\caption{Experimental results on RoadBEV under 10cm standard about the size of anchor cells $r$.}
\label{tab:roir}

\end{table}

\noindent\textbf{Size of anchor cells.} We also experiment on the size of anchor cells $r$, and the results on RoadBEV under 10 cm standard are in Tab.\ref{tab:roir}. With $r$ increasing, enlarged cells cover more room for local prediction to provide better refinement for potential imperfect global prediction. However, when $r$ is too large, it may cover multiple lines in one cell which sometimes leads to ambiguity, worsening the performance. In general, the shape of cells hardly affects results much so a fixed $r$ is enough for FHLD to get satisfying results.

\section{Conclusion}
In this paper, we propose an end-to-end framework FHLD which fully fuses information from global and local visions to output flexible and accurate results. In particular, lane line representations in different levels are proposed, and related hierarchical shape matching strategies are designed accordingly. Experiments on two datasets validate our superiority in improving local error under high accuracy requirements. Additionally, the idea of our work can be easily extended to 2D lane detection tasks and other inputs like multi-modal. 

\section{Acknowledgements}
This work was supported by the National Natural Science Foundation of China (61976170, 91648121, 62088102).

\nocite{li2021hdmapnet, yan2016scan, philion2019fastdraw, peng2022mass, RME}
\appendix
\bibliography{aaai23.bib}

\begin{thebibliography}{27}
\providecommand{\natexlab}[1]{#1}

\bibitem[{Chen et~al.(2020)Chen, Zhang, Zhong, Zhang, Ma, and Liu}]{dfpn}
Chen, S.; Zhang, Z.; Zhong, R.; Zhang, L.; Ma, H.; and Liu, L. 2020.
\newblock A dense feature pyramid network-based deep learning model for road
  marking instance segmentation using MLS point clouds.
\newblock \emph{IEEE Transactions on Geoscience and Remote Sensing}, 59(1):
  784--800.

\bibitem[{Chen, Liu, and Lian(2019)}]{pointlanenet}
Chen, Z.; Liu, Q.; and Lian, C. 2019.
\newblock Pointlanenet: Efficient end-to-end cnns for accurate real-time lane
  detection.
\newblock In \emph{2019 IEEE intelligent vehicles symposium (IV)}, 2563--2568.
  IEEE.

\bibitem[{Feng et~al.(2022)Feng, Guo, Tan, Xu, Wang, and Ma}]{bezierlanenet}
Feng, Z.; Guo, S.; Tan, X.; Xu, K.; Wang, M.; and Ma, L. 2022.
\newblock Rethinking efficient lane detection via curve modeling.
\newblock In \emph{Computer Vision and Pattern Recognition}.

\bibitem[{He et~al.(2017)He, Gkioxari, Doll{\'{a}}r, and Girshick}]{maskrcnn}
He, K.; Gkioxari, G.; Doll{\'{a}}r, P.; and Girshick, R.~B. 2017.
\newblock Mask {R-CNN}.
\newblock \emph{CoRR}, abs/1703.06870.

\bibitem[{He et~al.(2015)He, Zhang, Ren, and Sun}]{resnet}
He, K.; Zhang, X.; Ren, S.; and Sun, J. 2015.
\newblock Deep Residual Learning for Image Recognition.
\newblock \emph{CoRR}, abs/1512.03385.

\bibitem[{Homayounfar et~al.(2018)Homayounfar, Ma, Lakshmikanth, and
  Urtasun}]{hran}
Homayounfar, N.; Ma, W.-C.; Lakshmikanth, S.~K.; and Urtasun, R. 2018.
\newblock Hierarchical recurrent attention networks for structured online maps.
\newblock In \emph{Proceedings of the IEEE Conference on Computer Vision and
  Pattern Recognition}, 3417--3426.

\bibitem[{Homayounfar et~al.(2019)Homayounfar, Ma, Liang, Wu, Fan, and
  Urtasun}]{dagmapper}
Homayounfar, N.; Ma, W.-C.; Liang, J.; Wu, X.; Fan, J.; and Urtasun, R. 2019.
\newblock Dagmapper: Learning to map by discovering lane topology.
\newblock In \emph{Proceedings of the IEEE/CVF International Conference on
  Computer Vision}, 2911--2920.

\bibitem[{Hou et~al.(2019)Hou, Ma, Liu, and Loy}]{enetsad}
Hou, Y.; Ma, Z.; Liu, C.; and Loy, C.~C. 2019.
\newblock Learning lightweight lane detection cnns by self attention
  distillation.
\newblock In \emph{Proceedings of the IEEE/CVF international conference on
  computer vision}, 1013--1021.

\bibitem[{Jeong et~al.(2019)Jeong, Cho, Shin, Roh, and Kim}]{keist}
Jeong, J.; Cho, Y.; Shin, Y.-S.; Roh, H.; and Kim, A. 2019.
\newblock Complex Urban Dataset with Multi-level Sensors from Highly Diverse
  Urban Environments.
\newblock \emph{International Journal of Robotics Research}, 38(6): 642--657.

\bibitem[{Jung et~al.(2019)Jung, Che, Olsen, and Parrish}]{traditional1}
Jung, J.; Che, E.; Olsen, M.~J.; and Parrish, C. 2019.
\newblock Efficient and robust lane marking extraction from mobile lidar point
  clouds.
\newblock \emph{ISPRS journal of photogrammetry and remote sensing}, 147:
  1--18.

\bibitem[{Ko et~al.(2021)Ko, Lee, Azam, Munir, Jeon, and Pedrycz}]{pinet}
Ko, Y.; Lee, Y.; Azam, S.; Munir, F.; Jeon, M.; and Pedrycz, W. 2021.
\newblock Key points estimation and point instance segmentation approach for
  lane detection.
\newblock \emph{IEEE Transactions on Intelligent Transportation Systems}.

\bibitem[{Li et~al.(2021)Li, Wang, Wang, and Zhao}]{li2021hdmapnet}
Li, Q.; Wang, Y.; Wang, Y.; and Zhao, H. 2021.
\newblock Hdmapnet: A local semantic map learning and evaluation framework.
\newblock \emph{arXiv preprint arXiv:2107.06307}.

\bibitem[{Liu et~al.(2021{\natexlab{a}})Liu, Chen, Zhu, and Tan}]{condlanenet}
Liu, L.; Chen, X.; Zhu, S.; and Tan, P. 2021{\natexlab{a}}.
\newblock Condlanenet: a top-to-down lane detection framework based on
  conditional convolution.
\newblock In \emph{Proceedings of the IEEE/CVF International Conference on
  Computer Vision}, 3773--3782.

\bibitem[{Liu et~al.(2021{\natexlab{b}})Liu, Yuan, Liu, and Xiong}]{LSTR}
Liu, R.; Yuan, Z.; Liu, T.; and Xiong, Z. 2021{\natexlab{b}}.
\newblock End-to-end Lane Shape Prediction with Transformers.
\newblock In \emph{WACV}.

\bibitem[{Ma et~al.(2020)Ma, Li, Li, Yu, Junior, Gon{\c{c}}alves, and
  Chapman}]{unet2}
Ma, L.; Li, Y.; Li, J.; Yu, Y.; Junior, J.~M.; Gon{\c{c}}alves, W.~N.; and
  Chapman, M.~A. 2020.
\newblock Capsule-based networks for road marking extraction and classification
  from mobile LiDAR point clouds.
\newblock \emph{IEEE Transactions on Intelligent Transportation Systems},
  22(4): 1981--1995.

\bibitem[{Pan et~al.(2019)Pan, Yang, Li, Yang, Dong, and Yang}]{RME}
Pan, Y.; Yang, B.; Li, S.; Yang, H.; Dong, Z.; and Yang, X. 2019.
\newblock Automatic Road Markings Extraction, Classification and Vectorization
  from Mobile Laser Scanning Data.
\newblock \emph{International Archives of the Photogrammetry, Remote Sensing \&
  Spatial Information Sciences}.

\bibitem[{Peng et~al.(2022)Peng, Fei, Yang, Roitberg, Zhang, Bieder,
  Heidenreich, Stiller, and Stiefelhagen}]{peng2022mass}
Peng, K.; Fei, J.; Yang, K.; Roitberg, A.; Zhang, J.; Bieder, F.; Heidenreich,
  P.; Stiller, C.; and Stiefelhagen, R. 2022.
\newblock MASS: Multi-attentional semantic segmentation of LiDAR data for dense
  top-view understanding.
\newblock \emph{IEEE Transactions on Intelligent Transportation Systems}.

\bibitem[{Philion(2019)}]{philion2019fastdraw}
Philion, J. 2019.
\newblock Fastdraw: Addressing the long tail of lane detection by adapting a
  sequential prediction network.
\newblock In \emph{Proceedings of the IEEE/CVF Conference on Computer Vision
  and Pattern Recognition}, 11582--11591.

\bibitem[{Qin, Wang, and Li(2020)}]{ufast}
Qin, Z.; Wang, H.; and Li, X. 2020.
\newblock Ultra Fast Structure-aware Deep Lane Detection.
\newblock In \emph{The European Conference on Computer Vision (ECCV)}.

\bibitem[{Su et~al.(2021)Su, Chen, Zhang, Luo, Wei, and Wei}]{structure}
Su, J.; Chen, C.; Zhang, K.; Luo, J.; Wei, X.; and Wei, X. 2021.
\newblock Structure guided lane detection.
\newblock \emph{arXiv preprint arXiv:2105.05403}.

\bibitem[{Sun, Tsai, and Chan(2006)}]{HSI}
Sun, T.-Y.; Tsai, S.-J.; and Chan, V. 2006.
\newblock HSI color model based lane-marking detection.
\newblock In \emph{2006 ieee intelligent transportation systems conference},
  1168--1172. IEEE.

\bibitem[{Tabelini et~al.(2021{\natexlab{a}})Tabelini, Berriel, Paixao, Badue,
  De~Souza, and Oliveira-Santos}]{laneatt}
Tabelini, L.; Berriel, R.; Paixao, T.~M.; Badue, C.; De~Souza, A.~F.; and
  Oliveira-Santos, T. 2021{\natexlab{a}}.
\newblock Keep your eyes on the lane: Real-time attention-guided lane
  detection.
\newblock In \emph{Proceedings of the IEEE/CVF conference on computer vision
  and pattern recognition}, 294--302.

\bibitem[{Tabelini et~al.(2021{\natexlab{b}})Tabelini, Berriel, Paixao, Badue,
  De~Souza, and Oliveira-Santos}]{polylanenet}
Tabelini, L.; Berriel, R.; Paixao, T.~M.; Badue, C.; De~Souza, A.~F.; and
  Oliveira-Santos, T. 2021{\natexlab{b}}.
\newblock Polylanenet: Lane estimation via deep polynomial regression.
\newblock In \emph{2020 25th International Conference on Pattern Recognition
  (ICPR)}, 6150--6156. IEEE.

\bibitem[{Wen et~al.(2019)Wen, Sun, Li, Wang, Guo, and Habib}]{unet1}
Wen, C.; Sun, X.; Li, J.; Wang, C.; Guo, Y.; and Habib, A. 2019.
\newblock A deep learning framework for road marking extraction, classification
  and completion from mobile laser scanning point clouds.
\newblock \emph{ISPRS journal of photogrammetry and remote sensing}, 147:
  178--192.

\bibitem[{Yan et~al.(2016)Yan, Liu, Tan, Li, Xie, and Chen}]{yan2016scan}
Yan, L.; Liu, H.; Tan, J.; Li, Z.; Xie, H.; and Chen, C. 2016.
\newblock Scan line based road marking extraction from mobile LiDAR point
  clouds.
\newblock \emph{Sensors}, 16(6): 903.

\bibitem[{Yang et~al.(2021)Yang, Yang, Yang, Ming, Wang, Tian, and Yan}]{KLD}
Yang, X.; Yang, X.; Yang, J.; Ming, Q.; Wang, W.; Tian, Q.; and Yan, J. 2021.
\newblock Learning High-Precision Bounding Box for Rotated Object Detection via
  Kullback-Leibler Divergence.
\newblock \emph{CoRR}, abs/2106.01883.

\bibitem[{Zheng et~al.(2021)Zheng, Fang, Zhang, Tang, Yang, Liu, and
  Cai}]{resa}
Zheng, T.; Fang, H.; Zhang, Y.; Tang, W.; Yang, Z.; Liu, H.; and Cai, D. 2021.
\newblock Resa: Recurrent feature-shift aggregator for lane detection.
\newblock In \emph{Proceedings of the AAAI Conference on Artificial
  Intelligence}, volume~35, 3547--3554.

\end{thebibliography}
\end{document}